\documentclass[a4paper]{article}
\usepackage[numbers]{natbib}
\pdfoutput=1
\usepackage{pdfpages}

\usepackage{amsmath}
\usepackage{amsfonts}
\usepackage{amssymb}
\usepackage{graphicx}
\usepackage{multirow}
\usepackage[T1]{fontenc}

\newtheorem{thm}{Theorem}

\newtheorem{prf}{Proof}

\newcommand{\E}{\mathbb{E}}
\def \xx {\mathbf{x}}
\def \pp {\mathbf{p}}
\def \uu {\mathbf{u}}

\def \X  {\mathcal{X}}
\def \Y  {\mathcal{Y}}

\def \R  {\mathbb{R}}

\def \D  {\mathcal{D}}
\def \C  {\mathcal{C}}

\def \ej {\mathbf{e_j}}
\def \ei {\mathbf{e_i}}
\DeclareMathOperator*{\argmin}{arg\,min}
\DeclareMathOperator*{\argmax}{arg\,max}
\title{Robust Loss Functions under Label Noise for Deep Neural Networks}
\author{
	Aritra Ghosh\\
	Microsoft, Bangalore\\
	\texttt{arghosh@microsoft.com} \\
	  \and
	 Himanshu Kumar \\
	 Indian Institute of Science, Bangalore \\
	 \texttt{himanshukr@ee.iisc.ernet.in} \\
	 \and
	 P. S. Sastry \\
	 Indian Institute of Science, Bangalore \\
	 \texttt{sastry@ee.iisc.ernet.in} 
}
\date{}
\begin{document}
\maketitle
\begin{abstract}
In many applications of classifier learning, training data suffers from label noise. Deep networks are learned using huge training data where the problem of noisy labels is particularly relevant. The current techniques proposed for learning deep networks under label noise focus on modifying the network architecture and on algorithms for estimating true labels from noisy labels. An alternate approach would be to look for loss functions that are inherently noise-tolerant. For binary classification there exist theoretical results on loss functions that are robust to label noise. 
In this paper, we provide some sufficient conditions on a loss function so that risk minimization under that loss function would be inherently tolerant to label noise for multiclass classification problems. These results generalize the existing results on noise-tolerant loss functions for binary classification.  We study some of the widely used loss functions in deep networks  and show that the loss function based on mean absolute value of error is inherently robust to label noise. Thus standard back propagation is enough to learn the true classifier even under label noise. Through experiments, we illustrate the robustness of risk minimization with such loss functions for learning neural networks.
\end{abstract}

\section{Introduction}
Recently, deep neural networks have exhibited very impressive performance in many classification problems. However, in all such cases 
one needs very large training data. Labelling the training data patterns and ensuring correctness of the labels thus becomes a serious challenge in many applications of deep neural networks. 

When the class labels in the training data are noisy (i.e., may be incorrect) then it is referred to as label noise. Human labelling errors, measurement errors, subjective biases of labellers are among the reasons for label noise. In many large scale classification problems, labelled data is often obtained through crowd sourcing or by automatically using information on the web. This is another main reason for unreliability of labels in the training data. 

Robust learning of classifiers in the presence of label noise has been investigated from many viewpoints. In this paper, we study it in the framework of risk minimization which is a popular method for classifier learning. For example, Bayes classifier minimizes risk under 0--1 loss function. The standard backpropagation-based learning of neural networks is also risk minimization under different loss functions (such as squared error or cross entropy). The robustness of risk minimization depends on the loss function used. We call a loss function noise-tolerant if 
minimizer of risk (under that loss function) with noisy labels would be same as that with noise-free labels.

In this paper we present some novel analytical results on noise-tolerance of loss functions in a multiclass classification scenario. We derive sufficient conditions on a loss function so that it would be noise-tolerant for different types of label noise. 
We then examine some of the popular loss functions used for learning neural networks and show that loss function based on mean-absolute error~(MAE) satisfies our sufficient conditions. Empirical investigations are presented to compare robustness of learning neural networks under label noise using different loss functions based on mean-absolute error, mean-square error and categorical cross entropy. The empirical results well demonstrate the utility of the theoretical results presented here.

\subsection{Related Work}

Learning in presence of label noise is a long standing problem in machine learning.  
A detailed survey is available in \cite{frenay2013classification}.

There are many approaches to learning under label noise.
Data cleaning approaches rely on finding points which are corrupted by label noise. 
Once these points are identified, they can be either filtered out or their labels suitably altered. 
Several heuristics have been used to guess such noisy points \cite{Angelova:2005:PTS:1068507.1068955,Brodley99identifyingmislabeled,Zhu:2003}.
There have also been attempts at (heuristically) modifying existing learning algorithms to make them robust 
\cite{Khardon:2007:NTV:1248659.1248667,jin2003new,DBLP:journals/jmlr/BiggioNL11}.

Another prominent approach is to treat the (unknown) true labels as  hidden variables and to estimate a generative or discriminative model. \cite{lawrence2001estimating} proposed such a method, based on maximum-likelihood estimation of the model using the EM algorithm, in the context of fisher linear discriminant classifier. Similar methods have been proposed to make logistic regression robust to label noise \cite{bootkrajang2012label}. Recently many algorithms based on such ideas are proposed in deep learning literature to mitigate the adverse effect of label noise. \cite{mnih-hinton-2012} use a generative model for label corruption,  estimated through an approximate EM algorithm, and show its effectiveness in a binary classification problem. \cite{sukhbaatar2014training} proposed a modified architecture of a neural network to learn with noisy labels by effectively estimating label corruption probabilities.
Motivated by similar ideas,   a method to estimate a generative model incorporating label corruption is proposed in \cite{Xiao_2015_CVPR}. 
This method is applicable for multiclass classification and can handle fairly general cases of label noise. Methods based on  bootstrapping are also proposed for learning deep networks under label noise \cite{reed2014training}. While all these methods are seen to deliver good performance, they do not guarantee, in any probabilistic sense, robustness to label noise. 

All the above methods focus on changing the learning algorithm so that one can estimate the true labels of the training examples and thus be able to learn under label noise. As opposed to this, one can also look for methods that are inherently noise tolerant. Such algorithms treat noisy data and noise-free data the same way but achieve noise robustness due to properties of the algorithm. Such methods have been mostly investigated in the framework of risk minimization. 

Robustness of risk minimization depends on the loss function.
For binary classification, it is shown that 0--1 loss is robust to symmetric or uniform  label  noise while most of the standard convex loss functions are not robust \cite{long2010random,manwani2013noise}. The problem of learning from positive and unlabeled data can be cast as learning under label noise and it is seen that none of the common convex surrogate losses are good for this problem \cite{du2014analysis}. Unhinged loss, a convex loss (which is not convex potential), is robust to symmetric label noise \cite{van2015learning}.  \cite {natarajan2013learning} proposed a method for robust risk minimization through an implicit estimation of noise probabilities. In a similar spirit, \cite{DBLP:conf/colt/ScottBH13} proposed a method of estimating Type~{\rm 1} and Type~{\rm 2} error rates of any specific classifier under the noise-free distribution given only the noisy  training data. Recently, a general sufficient condition on a loss function is derived so that risk minimization is robust to label noise  \cite{ghosh2015making}. It is shown that the 0--1 loss, ramp loss and sigmoidal loss satisfy this condition. 

All the above are for the case of binary classification. Recently \cite{patrini2016making} proposed a robust risk minimization approach for learning neural networks for multiclass classification by estimating label corruption probabilities.  In our work here we also investigate robustness of risk minimization in the context of multiclass classification. We provide analytical results on conditions under which risk minimization is robust to different types of label noise. Our results generalize the existing results for 2-class problems. The currently known noise-tolerant loss functions (such as 0--1 loss or ramp loss) are not commonly used while learning neural networks. In this paper, we examine some common loss functions for learning neural networks for multiclass classification and show that the one based on mean absolute error is noise-tolerant. Through empirical studies we demonstrate the relevance of our theoretical results.

\section{Preliminaries and Problem Statement}

In this section we introduce some notation and 
define the notion of noise tolerance of a loss function.  

\subsection{Risk Minimization}
\label{sec:risk-minimization}
Let $\X \subset \R^d$ be the feature space from which
the examples are drawn and let $\Y=[k] = \{1,\cdots,k\}$ be the class labels.
In a typical classifier learning problem, we are given training data,
$S=\{(\xx_1,y_{\xx_1}),\ldots, (\xx_N,y_{\xx_N})\} \in (\X \times \Y)^N$, 
drawn {\em iid} according to an unknown distribution, $\mathcal{D}$,  over $\X \times \Y$. We represent a classifier as 
$h(\xx) = \mbox{pred}\circ f(\xx)$ where $f: \X \rightarrow \C, \; \C\subseteq\R^k$.
Here, $h$ (which predicts the class label given $f(\xx)$) maps $\X$ to $\Y$. Even though the final classification decision on a feature vector $\xx$ is $\mbox{pred}\circ f(\xx)$, we use the notation of calling $f$ itself as the classifier.

A loss function is a map $L: \C \times \Y \rightarrow \R^+$. Given any loss function, $L$, and 
a classifier, $f$, we define the {\em $L$-risk} of $f$ by 
\begin{equation}
R_{L}(f) =  \E_{\D}[ L(f(\xx), y_{\xx})] =  \E_{\xx, y_{\xx}} [ L(f(\xx), y_{\xx})]
\label{eq:l-risk}
\end{equation}
where, as a notation throughout this paper, the $\E$ denotes expectation and its subscript indicates the random variables or the distribution with respect to which the expectation is taken. 
Under risk minimization framework, the objective is to learn a classifier, $f$, which is a global minimizer of $R_L$. Note that the L-risk, $R_{L}$, depends on $L$, the loss function. When $L$ happens to be the 0--1 loss, $R_L$ would be the usual Bayes risk. 

\subsection{Noise Tolerance of Loss Functions}

When there is label noise, the learner does not have access to the clean training data (represented by $S$ above).
The noisy training data available to the learner is $S_{\eta} = \{(\xx_n,\hat{y}_{\xx_n}),n=1,\cdots,N\}$ where,
\begin{align*}
\hat{y}_{\xx_n}\,=\,\begin{cases}
y_{\xx_n}&  \text{with probability~}(1-\eta_{\xx_n})\\
j,\;\; j \in [k], \; j\neq y_{\xx_n} & \text{with probability~} \; \bar{\eta}_{\xx_n j}
\end{cases}
\end{align*}
Note that, for all $\xx$, conditioned on $y_{\xx}=i$, we have $\sum_{j \neq i} \bar{\eta}_{\xx j} = \eta_{\xx}$.

In general, for any $\xx$, its true label (that is, label under distribution $\mathcal{D}$) is denoted by the random variable  $y_{\xx}$ while the 
noise corrupted label is denoted by $\hat{y}_{\xx}$. We use $\mathcal{D}_{\eta}$ to denote the 
joint probability distribution of $\xx$ and $\hat{y}_{\xx}$.

The noise is termed {\em symmetric} or {\em uniform} if $\eta_{\xx}=\eta, \mbox{~~and~~} \bar{\eta}_{\xx j}=\frac{\eta}{k-1}, \; \forall j\neq y_{\xx}, \forall \xx$, where $\eta$ is a constant. 

Noise is said to be {\em class-conditional} or {asymmetric} if the dependence of $\eta_{\xx}$ on $\xx$ is only through $y_{\xx}$ and similarly for $\bar{\eta}_{\xx j}$. 
In this case, with a little abuse of notation, we write $\eta_{\xx}=\eta_{y_{\xx}}, \; \bar{\eta}_{\xx j}=\bar{\eta}_{y_{\xx}j}$. Thus, for example, $\bar{\eta}_{ij}$ would be the probability that a class-$i$ pattern would have label as class-$j$ when the label is corrupted. 

In general, when noise rate $\eta_{\xx}$ as well as $\bar{\eta}_{\xx j}$ is a function of $\xx$, 
it is termed as {\em non-uniform} noise. A simple special case is when $\bar{\eta}_{\xx j}=\frac{\eta_{\xx}}{k-1}, \; \forall j\neq y_{\xx}$. We define it as {\em simple non-uniform noise}. Furthermore, when $\eta_{\xx}$ is fixed for each class, we call it {\em simple class conditional noise}.

The $L$-risk, $R_L(f)$, given by eq.\ref{eq:l-risk} is for the  noise-free case.
Let $f^*$ be the global minimizer (over the chosen function class) of $R_L(f)$. 
When there is label noise, the data is drawn according to distribution  $\mathcal{D}_{\eta}$. 
Then L-risk of a classifier $f$ under noisy data is
\[R_L^\eta(f)= \E_{\D_{\eta}}[L(f(\xx),\hat{y}_{\xx})] =  \E_{\xx, \hat{y}_{\xx}}[L(f(\xx),\hat{y}_{\xx})]\]
(We use $\E_{\xx, y_{\xx}}(\E_{\xx, \hat{y}_{\xx}})$ and $\E_{\D}(\E_{\D_{\eta}})$ interchangeably). Let $f^*_{\eta}$ be the global minimizer (over the chosen function class) of $R_L^\eta(f)$. 
Risk minimization under loss function $L$, is said to be {\em noise-tolerant} if~\cite{manwani2013noise}
\[{\Pr}_{\mathcal{D}}[\mbox{pred}\circ f^*(\xx)=y_{\xx}]={\Pr}_{\mathcal{D}}
[\mbox{pred}\circ f^*_{\eta}(\xx)=y_{\xx}]\]
Risk minimization under a given loss function is  
noise tolerant if the $f^*_{\eta}$ has the same probability of misclassification as
that of $f^*$  on the noise free data.
When the above is satisfied we also say that the loss function $L$ is noise-tolerant. For this, it is sufficient if $f^*=f_{\eta}^*$. 
\section{Theoretical Results}
We call a loss function $L$ symmetric if it  satisfies, for some constant $C$, 
\begin{equation}
\label{eq:symmetry}
\sum_{i=1}^{k} L(f(\xx),i)=C, \  \forall \xx \in \X, \forall f. 
\end{equation}
In the following, we prove distribution independent sufficient conditions for loss function to be robust under different kinds of label noises. In Theorem~1, we prove sufficiency results for {\em symmetric label noise}, followed by {\em simple non-uniform noise} and {\em class-conditional noise} in Theorems~2, 3.
\begin{thm}
	\label{th:uniform}	
	In a multi-class classification problem, let loss function $L$ satisfy Eq~\ref{eq:symmetry}.
	Then $L$ is noise tolerant under symmetric or uniform label noise if $\eta < \frac{k-1}{k}$.
\end{thm} 
\begin{prf}
	Recall that for any $f$, 
	\begin{equation}
	R_L(f)=\E_{\xx, y_{\xx}} L(f(\xx),y_{\xx})
	\end{equation}
	For uniform noise, we have, for any $f$,\footnote{In the following, $\E_{y_{\xx} | \xx} \E_{\hat{y}_{\xx} | \xx, y_{\xx}}$ etc. denote expectation with respect to the corresponding conditional distributions. Note that $\E_{\xx} \E_{ y_{\xx} | \xx}=\E_{\xx, y_{\xx}}=\E_{\D}$.}
	\begin{align*}
		R_L^\eta(f)&=  \E_{\xx, \hat{y}_{\xx}} L(f(\xx), \hat{y}_{\xx})\\
		&=  \E_{\xx} \E_{y_{\xx} | \xx} \E_{\hat{y}_{\xx} | \xx, y_{\xx}} L(f(\xx), \hat{y}_{\xx}) \\
		&= \E_{\xx} \E_{y_{\xx} | \xx} \left[ (1-\eta) L(f(\xx),y_{\xx}) + \frac{\eta}{k-1} \sum_{i\neq y_{\xx}} L(f(\xx),i) \right] \\
		& =  (1 - \eta) R_L(f) +  \frac{\eta}{k-1} (C - R_L(f))\\ 
		&= \frac{C \eta}{k-1} + \left(1-\frac{\eta k}{k-1}\right) R_L(f).		
	\end{align*}
	Thus, for any $f$,
	\[R_L^\eta(f^*)-R_L^\eta(f)=(1-\frac{\eta k}{k-1})(R_L(f^*)-R_L(f)) \leq 0\]
	because 
$\eta < \frac{k-1}{k}$  and $f^*$ is a minimizer of $R_L$.
	This proves $f^* $ is also minimizer of risk under uniform noise.
\end{prf}
\textbf{Remark 1 }Theorem~\ref{th:uniform} shows that symmetric losses are robust to uniform label noise. This does not depend on the data distribution. The only condition is that noise rate is less than $\frac{k-1}{k}$ which is not restrictive. 
	This theorem (along with the next one) generalizes the existing results for the 2-class case~\cite[Theorem~1]{ghosh2015making}.

\begin{thm}
	\label{th:nonuniform}
	Suppose loss $L$ satisfies Eq~\ref{eq:symmetry}. If $R_L(f^*)=0$, then $L$ is also noise tolerant under simple non uniform noise when $\eta_{\xx} < \frac{k-1}{k}$, $\forall \xx$.\\
	If $R_L(f^*) = \rho >0$ then, under simple non-uniform noise, $R_L(f^*_{\eta})$ is upper bounded by $\rho/(1-\frac{k\eta_{max}}{k-1})$, where $\eta_{max}$ is maximum noise rate over $\xx \in \X$. (Recall that $f^*$ is minimizer of $R_L$ and $f^*_{\eta}$ is minimizer of $R^{\eta}_L$). 
	
\end{thm}
\begin{prf}
	Under simple non-uniform noise, for any $f$, 
	\begin{equation*}
	\begin{split}
	R_L^\eta(f) =\; & \E_{\xx} \E_{y_{\xx} | \xx} \E_{\hat{y}_{\xx} | \xx, y_{\xx}} L(f(\xx), \hat{y}_{\xx})\\ 
        =\; &\E_{\D }\left[(1- \eta_{\xx}) L(f(\xx),y_{\xx})  
	+ \sum_{i\neq y_{\xx}} \frac{\eta_{\xx} L(f(\xx), i)}{k-1}\right] \\
	=\; &\E_{\D} (1- \eta_{\xx}) L(f(\xx),y_{\xx})+ \E_{\D} \frac{\eta_{\xx} }{k-1}\Big(C- L(f(\xx),y_{\xx})\Big)\\
	=\; &\E_{\D} C'\eta_{\xx} +\E_{\D}  \left((1-\frac{k\eta_{\xx}}{k-1})L(f(\xx),y_{\xx})\right)
	\end{split}
	\end{equation*}
	where $C'= \frac{C}{k-1}$. Hence we have
	\begin{equation}
	\label{eq:non_2}
	\begin{split}
	R_L^\eta(f^*)-R_L^\eta(f)= \E_{\D} \big\{ (1-\frac{k\eta_{\xx}}{k-1})(L(f^*(\xx),y_{\xx})-L(f(\xx),y_{\xx}))\big\}\\
	\end{split}
	\end{equation}
	Since $R_L(f^*) = 0$ and $L$ is non-negative by definition,
we have $L(f^*(\xx),y_{\xx})=0, \; \forall \xx$. 
	In addition, since $(1-\frac{k\eta_{\xx}}{k-1}) > 0$, we have $R_L^\eta(f^*)-R_L^\eta(f) \leq 0$, for any $f$. Thus minimizer of noise free case is also a minimizer of noisy case. This completes proof of first part of theorem.
	
	For the second part of the theorem, we have, 
	\begin{align}
	\label{eq:non_3}
	\begin{split}
	R_L^\eta(f^{\ast}_{\eta})-R_L^\eta(f^{\ast}) \leq 0 \\
	\Rightarrow\E_{\D}(1-\frac{k\eta_{\xx}}{k-1}) (L(f^*_{\eta}(\xx),y_{\xx})-
	L(f^{\ast}(\xx),y_{\xx}))\leq 0 \\
	\Rightarrow\min_{\eta_{\xx}} (1-\frac{k\eta_{\xx}}{k-1})\E_{\D} L(f^*_{\eta}(\xx),y_{\xx})\leq \E_{\D}L(f^{\ast}(\xx),y_{\xx})\\
         \Rightarrow R_L(f^{\ast}_{\eta})\leq \rho/(1-\frac{k\eta_{max}}{k-1})		
	\end{split}
	\end{align}
	where $\E_{\D}L(f^{\ast}(\xx),y_{\xx}) = R_L(f^*) = \rho$. 
Note that, in the above, we used  $\eta_{\xx}<\frac{k-1}{k}$, and hence  $0 < (1-\frac{k\eta_{\xx}}{k-1}) \leq 1, \; \forall \xx$. This completes the proof.
\end{prf}
	\textbf{Remark 2 }Theorem~\ref{th:nonuniform} establishes a sufficient condition for risk minimization to be robust to simple non uniform label noise. The condition needs $R_L(f^*)$ to be zero. If $L$ is the 0--1 loss then $R_L$ is the Bayes risk and then the sufficient condition is that the classes are separable (under noise-free case). However, even if the classes are separable, for a general loss function (e.g., sigmoidal loss), $R_L(f^*)$ may not be zero. The second part of Theorem~\ref{th:nonuniform} gives a bound on $R_L(f_{\eta}^*)$ in such cases. This part is useful even when classes are not separable and the optimal Bayes risk is non-zero. In case of high noise rate the bound might be loose, but one should note that, this is a distribution independent bound.
In the binary classification case, 	
	if data is separable, robustness can be achieved even though minimum value of L-risk is not zero if the loss function is `sufficiently steep' \cite[Theorems 2, 4]{ghosh2015making}. It appears possible to prove a similar result in multiclass case also.

\begin{thm}
	\label{th:class}
	Suppose $L$ satisfies Eq~\ref{eq:symmetry} and $0\leq L(f(\xx),i)\leq C/(k-1),\forall i\in [k]$. If $R_L(f^*)=0$, then, $L$ is noise tolerant under class conditional noise when $\bar{\eta}_{ij}<(1-\eta_i), \forall j\neq i$, $\forall i,j\in [k]$.	
\end{thm}
\begin{prf}
	For class-conditional noise, we have
	\small{
	\begin{align}
	\label{eq:cc_1}
	\begin{split}
	R_L^\eta(f)=\;  &\E_{\D} (1- \eta_{y_{\xx}}) L(f(\xx),y_{\xx})+  \E_{\D} \sum_{i\neq y_{\xx}} \bar{\eta}_{y_{\xx} i} L(f(\xx),i) \\
	= \;& \E_{\D} (1-\eta_{y_{\xx}})(C-\sum_{i\neq y_x}L(f(\xx),i))\E_{\D} \sum_{i\neq y_x} \bar{\eta}_{y_{\xx} i} L(f(x),i) \\
	=& C\E_{\D} (1-\eta_{y_{\xx}})-\E_{\D} \sum_{i\neq y_{\xx}}(1-\eta_{y_{\xx}}-\bar{\eta}_{y_{\xx} i}) L(f(\xx),i)
	\end{split}
	\end{align}	}
	Since $f^{\ast}_{\eta}$ is the minimizer of $R_L^\eta$, we have $R_L^{\eta}(f_{\eta}^{\ast})-R_L^{\eta}(f^{\ast})\leq  0$ and hence from Eq.(\ref{eq:cc_1}) we have
	
	\begin{align}
	\label{eq:cc_2}
	\E_{\D}\sum_{i\neq y_{\xx}}(1-\eta_{y_{\xx}}-\bar{\eta}_{y_{\xx} i})(L(f^{\ast}(\xx),i)-L(f^{\ast}_{\eta}(\xx),i)) \leq 0			
	\end{align}
	
	Since we are given $R_L(f^{\ast})=0$, we have $L(f^{\ast}(\xx),y_{\xx}) = 0$. Given the condition on $L$ in the theorem, this implies $L(f^{\ast}(\xx),i)=C/(k-1),\; i\neq y_{\xx}$. As per the assumption on noise in the theorem,  $(1-\eta_{y_{\xx}}-\bar{\eta}_{y_{\xx} i})>0$. Also, $L$ has to satisfy $L(f^{\ast}_{\eta}(\xx),i)\leq C/(k-1),\; \forall i$. Thus for Eq.(\ref{eq:cc_2}) to hold, it must be the case that $L(f^{\ast}_{\eta}(\xx),i)=C/(k-1), \; \forall i \neq y_{\xx}$ which, by symmetry of $L$, implies $L(f^{\ast}_{\eta}(\xx),y_{\xx}) = 0$. 
	Thus minimizer of true risk is also a minimizer of risk under noisy data. This completes the proof.
\end{prf}
	 \textbf{Remark 3 }Note that $1-\eta_{y_{\xx}}>\bar{\eta}_{y_{\xx} i}$ implies $\eta_{y_{\xx}}<(k-1)/k$. Thus the condition on noise rates for this theorem is more strict.  For $i\neq j$, $\bar{\eta}_{ij}$ is the probability that a feature vector of class-$i$ is labelled as class-$j$. If we set $\bar{\eta}_{ii} = 1- \eta_i$ which is the probability of a feature vector of class-$i$ having correct label, then the condition is that the matrix $[\bar{\eta}_{ij}]$ of label noise probabilities should be diagonal dominant. The condition on the loss function in the theorem is satisfied by some of the symmetric losses such as 0--1 loss and MAE loss. The condition that $R_L(f^*)=0$ is restrictive.   However, experimentally, even though minimum risk might not be $0$, symmetric losses show good robustness even under class-conditional noise.

\begin{table*}[t]	
	\centering
		\caption{Standard Datasets and Architecture}
	\resizebox{\columnwidth}{!}{

	\begin{tabular}{|c|c|}		
		\hline
		{Dataset} ($n_{tr},n_{te},c,d$) & {Hidden layer Architecture}\\
		\hline
		\multirow{2}{*}{MNIST ($60k,10k,10,28\times 28$)}& Conv layer + max pooling ($dr=0.25$)+  \\&two layers 1024 units (dr=0.25, 0.5)\\
		\hline
		\multirow{3}{*}{CIFAR 10 ($50k,10k,10,3\times 32\times 32$) \cite{krizhevsky2009learning}} &2 Conv layers + max pooling (dr=0.2)+ \\
		 
		&2 Conv layers + max-pooling (dr=0.2)+ \\
		&1 layer 512 units (dr=0.5) \\ \hline 
		Reuters RCV1 ($213k,213k,50,2000$)\cite{lewis2004rcv1}&One layer 256 units (dr=0.5)\\ 
		\hline 
		Reuters newswire ($8982,2246,46,2k$) & One layer 128 units (dr=0.5)\\ \hline
		\multirow{2}{*}{20 newsgroup by-date ($11314,7532,20,5k$)} & Input directly connected to Softmax layer\\ & with max-norm constraint\\ \hline
		\multirow{2}{*}{Imdb Sentiment ($20k,5k,2,5k$) \cite{maas-EtAl:2011:ACL-HLT2011}}& One embedding layer 50 units (dr=0.2)+\\
		& Conv layer + one layer 250 units (dr=0.5)\\ \hline					
	\end{tabular} 
	\label{datasets}}
\end{table*}
\subsection{Some Loss Functions for Neural Networks}
We assume standard neural network architecture with softmax output layer. 
If input to network is $\xx$, then input to softmax layer is $f(\xx)$. 
Softmax layer  computes:
\[u_i=\frac{\exp(f(\xx)_i)}{\sum_{j=1}^k \exp(f(\xx)_j)},\quad i\in [k]\]
where  $f(\xx)_i$ represents $i^{\mbox{th}}$ component of $f(\xx)$.
We have $\sum_{i=1}^{k}u_i=1$.  We define $\uu=[u_1,\cdots,u_k]$. 
The label for the training patterns is in `one-of-K' representation. If the class of $\xx$ is $j$ then $y_{\xx}$ is given as $\ej$ where
$e_{ji}=1$ if $i= j$, otherwise $0$.
We can now define some popular loss functions  namely,  categorical cross entropy (CCE), Mean square error (MSE) and Mean absolute error (MAE) as below.

\begin{align*}
L (f(\xx),\ej)\,=\begin{cases}
\sum_{i=1}^{k} e_{ji}\log\frac{1}{u_i} =\log\frac{1}{u_j} & \text{CCE}\\
||\ej-u||_1
=2-2u_j& \text{MAE}\\
||\ej-u||_2^2
= ||u||_2^2+1-2u_j&\text{MSE}
\end{cases}
\end{align*}
For these loss functions, we have
\begin{align*}
\sum_{i=1}^k L(f(\xx), \ei)\,=\,\begin{cases}
\sum_{i=1}^{k} \log\frac{1}{u_i} & \text{CCE}\\
\sum_{i=1}^{k} (2-2u_i)=2k-2& \text{MAE}\\
k||u||_2^2+k-2& \text{MSE}
\end{cases}
\end{align*}
Thus, among these, only MAE satisfies symmetry condition given by Eq.(\ref{eq:symmetry}). 
While MSE does not satisfy Eq.(\ref{eq:symmetry}), one can show, using $\frac{1}{k}\leq ||u||_2^2\leq1$, that
$k-1\leq\sum_{i}^{k} L_{mse}(f(x),i)\leq 2k-2$.
This boundedness makes it more robust than an unbounded loss such as CCE. 

Informally, a loss function is said to be classification calibrated if a classifier having low enough risk under that loss would also have low risk under 0--1 loss. 
Logistic loss and exponential loss in multi-class settings have been proved to be classification calibrated \cite{Weston98multi-classsupport,tewari2007consistency,bartlett2006convexity}. One can show that, MAE, MSE losses are also classification calibrated. 

\section{Classification Calibrated Losses}
For the sake of completeness, we show that  loss functions considered in the main paper are classification calibrated. We follow the notation and convention defined in \cite{bartlett2006convexity,tewari2007consistency}. We denote $L_{\psi}$ and $R_{\psi}$ as the conditional $\psi$ risk and conditional $\psi$ regret. 
In multi-class case, a surrogate loss function $\psi:[k]\times \C \rightarrow \R,\quad \C \subseteq \R^k$, is called $0-1$ classification calibrated if, $\forall \pp\in \Delta^k$,
\begin{equation}
\label{eq:calibrated_loss}
\begin{split}
\inf_{\uu\in \C:pred(\uu)\notin \argmin_{t\in [k]}L_{0-1}(\pp,t)}L_{\psi}(\pp,\uu)>&\inf_{u \in \C} L_{\psi}(\pp,\uu)\\
\inf_{\uu\in \C: R_{0-1}(\pp,pred(\uu))>0}R_{\psi}(\pp,\uu)>& 0
\end{split}
\end{equation}
Logistic loss and exponential loss in multi-class settings have been proved to be classification calibrated. \cite{Weston98multi-classsupport,tewari2007consistency}.
We show MAE and MSE are also classification calibrated. Note that, $\sum_{i=1}^k u_i=1$ for both these losses. 

We define $S_{\pp}=\{\uu\in \C: R_{0-1}(\pp,pred(\uu)>0)\}$. It is easy to see that $S_{\pp}=\{\uu \in \C:\argmax_i u_i\neq \argmax_i p_i \}$.
For MAE ,
\begin{equation}
\begin{split}
\inf_{\uu\in \C} L_{\psi}(\pp,\uu)=\inf_{\uu\in \C} \sum_{i=1}^k |p_i-u_i|
\end{split}
\end{equation}
For MSE,
\begin{equation}
\begin{split}
\inf_{\uu\in \C} L_{\psi}(\pp,\uu)=\inf_{\uu\in \C} \sum_{i=1}^k (p_i-u_i)^2
\end{split}
\end{equation}
For both these cases, unique minimizer is $p_i=u_i$. The $S_{\pp}$ is set is non-empty unless $p_i=1/k\quad \forall k$.
Thus the conditional $\psi$ regret is always positive on the set $S_{\pp}$. This completes the proof.
\subsection{Consistency under Symmetric Label Noise}
We showed that risk minimization with symmetric losses is robust to label noise. Since there is only a finite training set, one can only minimize Empirical Risk. 
We now prove consistency of empirical risk minimization under label noise.
\begin{thm}
	\label{vc}
	Consider empirical risk minimization (ERM) under symmetric label noise over a given function class of finite VC dimension.  If the loss $L$ used for ERM is robust to label noise (i.e., satisfies eq.~\ref{eq:symmetry}), then the error rate of minimizer of empirical risk with noisy samples converges uniformly  to the error rate of the minimizer of risk under noise-free distribution. 
\end{thm}
\begin{prf}
	\hfill
	
	We denote by $er_{\D}[g]$ the error rate (i.e., 0-1 risk) of classifier $g$ in the noise-free case.  Under noise we denote it as $er_{\D_{\eta}}[g]$. 
	Let $\hat{g}^{\ast}$ ($\hat{g}^{\ast}_{\eta}$) be the minimizer of empirical risk over $n$ noise-free (noisy) samples. Let $g^{\ast}$ ($g^{\ast}_{\eta}$) be the minimizer of risk. Since VC bounds are distribution independent, we have,
	\begin{flalign*}
		er_{\D}[\hat{g}^{\ast}]\leq& er_{\D}(g^{\ast})+\epsilon(n, vc)\\
		er_{\D_{\eta}}[\hat{g}_{\eta}^{\ast}]\leq& er_{\D_{\eta}}(g_{\eta}^{\ast})+\epsilon(n, vc) 
	\end{flalign*}

	The term $\epsilon(n,vc)$ or simply $\epsilon$ goes to $0$ with $\frac{vc}{n}$, where $vc$ is the VC dimension of the function class  \cite{Vapnik:1995:NSL:211359}.
	
	Under symmetric label noise $\eta$, we derived how error rate changes. (Note that 0-1 loss is symmetric). Thus,
	\[er_{\D_{\eta}}=er_{\D}(1-\frac{k\eta}{k-1})+c'\eta\]
	Then we have the following,
	\begin{flalign*}
	&er_{\D_{\eta}}[\hat{g}^{\ast}_{\eta}]-er_{\D_{\eta}}[g^{\ast}_{\eta}]
	=(er_{\D}[\hat{g}^{\ast}_{\eta}]-er_{\D}[g_{\eta}^{\ast}])(1-\frac{k\eta}{k-1})\leq \epsilon\\
	&\Rightarrow\  	er_{\D}[\hat{g}^{\ast}_{\eta}]-er_{\D}[g^{\ast}_{\eta}]=
	er_{\D}[\hat{g}^{\ast}_{\eta}]-er_{\D}[g^{\ast}]\leq \frac{\epsilon}{1-\frac{k\eta}{k-1}}	
	\end{flalign*}
	where we have used $er_{\D}[g^{\ast}_{\eta}]=er_{\D}[g^{\ast}]$ which follows because $L$ is robust to label noise.
	This completes the proof.	
\end{prf}

	\section{Empirical Results}
	
	In this section we illustrate the robustness of symmetric loss functions. We present results with two image data sets and four text data sets. In each case we learn a neural network classifier using the CCE, MSE and MAE loss functions. We add symmetric or class conditional noise with different noise rates to the training set. For learning, we minimize the empirical risk, with different loss functions, using stochastic gradient descent through backpropagation \cite{bergstra2010theano,chollet2015keras}. The learnt networks are tested on noise-free test sets.

	\begin{figure*}[t]
		\centering
		\resizebox{\columnwidth}{!}{
			\begin{tabular}{cccc}
				\includegraphics[scale=0.6]{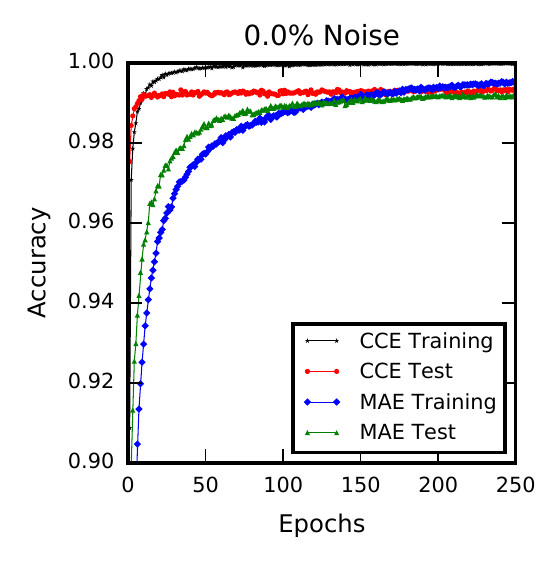}  &  \includegraphics[scale=0.6]{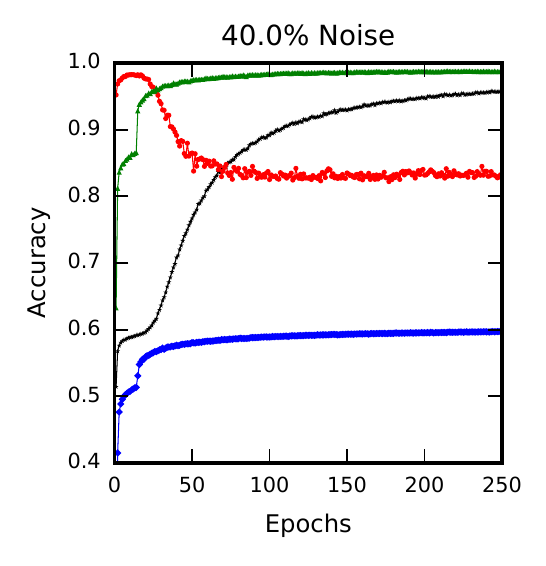} & \includegraphics[scale=0.6]{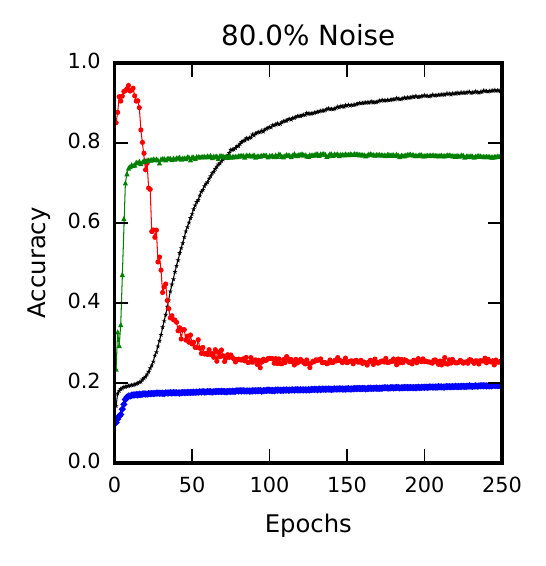}  &  \includegraphics[scale=0.6]{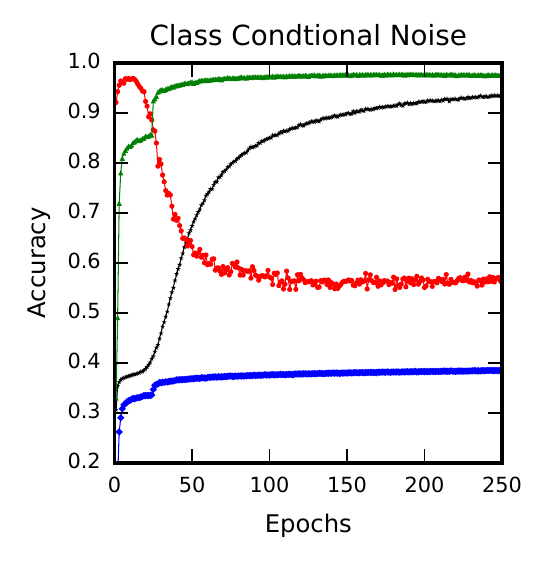}\\
				(a) & (b) & (c) & (d)\\
				\includegraphics[scale=0.6]{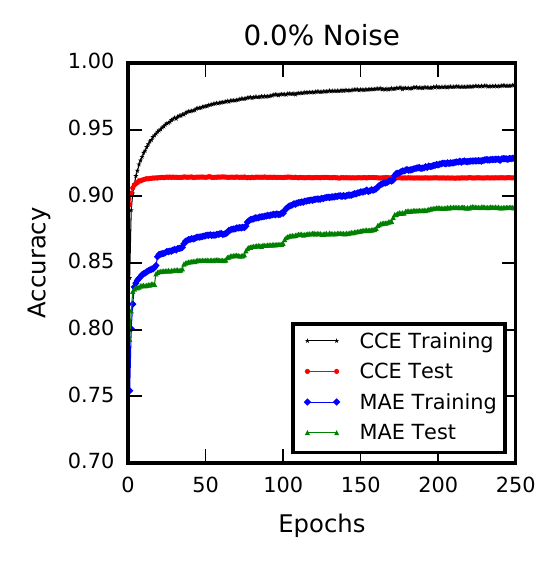}  &  \includegraphics[scale=0.6]{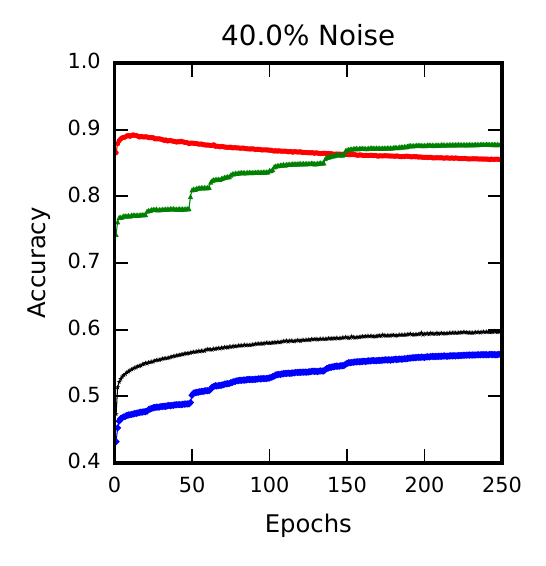} & \includegraphics[scale=0.6]{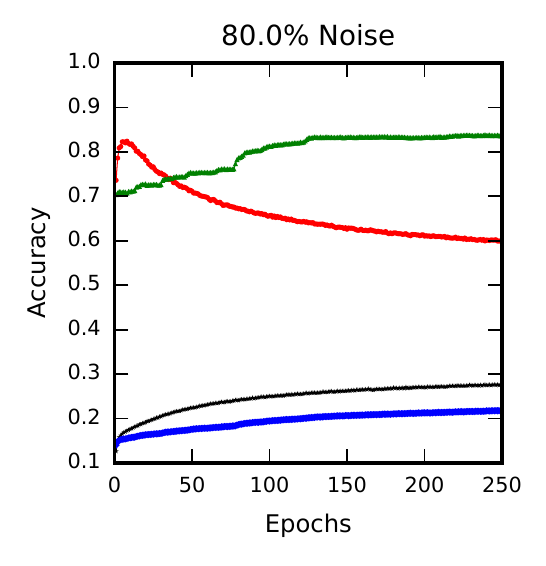}  &  \includegraphics[scale=0.6]{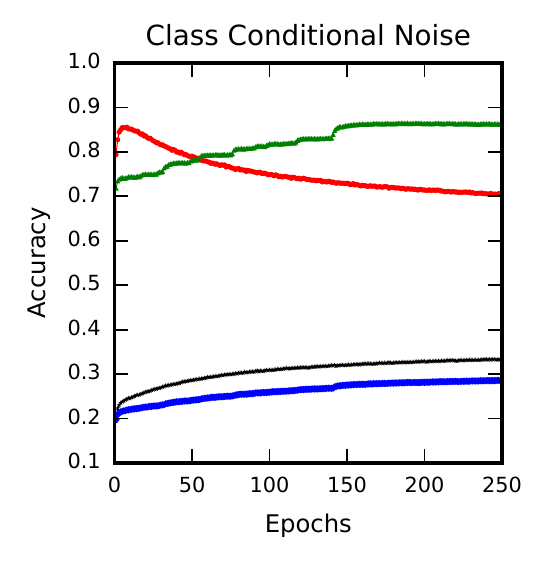}\\
				(e) & (f) & (g) & (h)\\
				
			\end{tabular}}
			\caption{\footnotesize{Train-Test Accuracies for log loss and MAE over epochs, for MNIST Datasets under (a) $0\%$ noise (b) $40\%$ noise (c) $80\%$ noise (d) CC noise; and
					RCV1 Datasets under (e) $0\%$ noise (f) $40\%$ noise (g) $80\%$ noise (h) CC noise. Legends shown in (a) and (e).}}
			\label{fig:epoch}
	\end{figure*}

	\subsection{Experimental Setting}
	The specific image and text data sets used are shown in Table~\ref{datasets}. In the table, for each data set, we mention size of training and test sets ($n_{tr},n_{te}$), number of classes ($c$) and input dimension ($d$). Since some are image data while others are text data and feature space dimensions are all different, we have used different network architectures for each data set. These are also specified in Table~\ref{datasets}. All networks used Rectified Linear Unit (ReLU) in the hidden layers and have softmax layer at the output with the size of the layer being the number of classes. All networks are trained through backpropagation with momentum term and weight decay. We have also used dropout regularization and the dropout rates are also shown in Table~\ref{datasets}. 

The results reported are averages over six runs. Label noise is added in the training set by changing the label of each example independently. For symmetric noise, we fix $\eta$ and randomly change the label of each example. For class conditional noise, for each experiment, we generate a fixed label noise probability matrix, $[\bar{\eta}_{ij}]$, randomly and use that to decide the new labels. We ensure that the matrix is diagonal dominant as needed for our theoretical results.

	\begin{table}
		\caption{Accuracies under different noise rates ($\eta$) for all datasets (for Imdb, $\eta$'s are halved). The last column gives accuracies under class conditional noise. In all the cases, standard deviation is shown only when it is more than $0.01$}
		\resizebox{\columnwidth}{!}{
			\begin{tabular}{|c|c|c|c|c|c|}
				\hline
				\hline

				Data & loss & $\eta=0\%$ & $\eta=30\%$  & $\eta=60\%$  & CC \\ \hline
				
				{\parbox{1cm}{MNIST}} & {CCE} & $0.9936$   & $0.9138 $   & $0.5888 $ & $0.5775 (\pm 0.0291)$\\
				& MAE & $0.9916 $   & $0.9886$   & $0.9799$ & $0.9713 $ \\
				& MSE & $0.9921 $   & $0.9868 $   & $0.9766 $ & $0.8505(\pm 0.0473)$ \\ \hline
				
				{\parbox{1cm}{RCV1}} &CCE& $0.9126 $   & $0.8738 $   & $0.7905 $ & $0.7418 (\pm 0.025)$ \\
				& MAE& $0.8732(\pm 0.0107) $   & $0.8688 $   & $0.8637 (\pm 0.0201)$ & $0.8587$\\
				& MSE & $0.9014 $  & $0.8943 $   & $0.8682(\pm 0.0120)$ &$ 0.8315$ \\ \hline
				
				{\parbox{1cm}{Cifar 10}} &CCE & $0.7812 $ & $0.5598 (\pm 0.0170) $  & $0.3083 $ & $0.4896 $\\
				& MAE & $0.7810(\pm 0.0190) $  & $0.7011 (\pm 0.0264) $ & $0.5328(\pm 0.0251) $ & $0.61425(\pm 0.0320) $ \\
				& MSE & $0.8074 $  & $0.7027 $ & $0.5257(\pm 0.0146) $ & $0.6249(\pm 0.0359) $ \\
				\hline

				{\parbox{1cm}{Imdb}} & CCE & $0.8808 $   & $0.7729 $   & $0.6466 $ & $0.7858 (\pm 0.0135) $  \\
				& MAE & $0.8813 $   & $0.8500 $   & $0.7352(\pm 0.0145)  $ & $0.8382(\pm 0.0127)  $ \\
				& MSE  & $0.8816 $   & $0.7725(\pm 0.0105)  $   & $0.6506 (\pm 0.0103) $ & $0.7874 $  \\
				\hline
				
				{\parbox{1cm}{News wire}} & {CCE} & $0.7842  $   & $0.6905 $   & $0.4670 $ & $0.4973 (\pm 0.0148)$\\
				&MAE& $0.8081 $   & $0.7553 $   & $0.6357(\pm 0.0106)  $ & $0.6535 $ \\
				& {MSE} & $0.7916$   & $0.6626 $   & $0.4078(\pm 0.0172)  $ & $0.4377(\pm 0.0140) $ \\
				\hline
				{\parbox{1cm}{News group}} & CCE & $0.8006 $   & $0.7571 $   & $0.6435 $ & $0.5629 $  \\
				& MAE & $0.7890 $   & $0.7749 $   & $0.7319 $ & $0.6772 $ \\
				& MSE  & $0.7999 $   & $0.7553 $   & $0.6347 $ & $0.5519 $  \\ \hline
				
			\end{tabular}
			}
			\label{tab_acc}
			
		\end{table}	
	\subsection{Results and Discussion}
	
	In figure~\ref{fig:epoch},  we compare the robustness of MAE and CCE losses on MNIST image data set and RCV1 text data set. We have used symmetric label noise with $\eta = 0, \; 0.4, \; 0.8$. The figure  shows the evolution of training and test accuracies of the network with number of epochs of training. As the graphs in Fig.~\ref{fig:epoch}(a)--(c) show, MAE loss is highly robust to symmetric label noise. The test accuracy achieved with MAE even under 80\% noise is close to that with zero noise. On noise-free data, the accuracy achieved with CCE loss is a little bit higher than that with MAE. However, even at 40\% noise, the test accuracy with CCE loss drops sharply. Similar trend can be seen on the RCV1 data. (See Fig.~\ref{fig:epoch}(e)--(g)). Here, eventhough the drop in accuracy of CCE loss is not as sharp, it is clearly seen that MAE is much more robust. 
	In Fig.~\ref{fig:epoch} (d) and (h) we show results under class-conditional noise (CC).  
In these problems we have no idea whether the minimum risk is zero. However, as can be seen from the figure, the symmetric MAE loss exhibits a good level of robustness under class conditional noise also. 
	
	Table~\ref{tab_acc}, shows average test accuracy  and standard deviation (over six runs)  of the learnt networks for different noise rates. 
	We show results for noise rate $\eta =0.0,\;0.3,\;0.6$. (For the 2-class Imdb dataset,  noise rate used is $\eta =0.0,\;0.15,\;0.3$). We also show results for class conditional noise. As can be seen from the table, MAE exhibits  good robustness. When the accuracy of MAE at 0\% noise is high (e.g., MNIST and RCV1), it drops very little even under 60\% noise rate. However, when accuracy achieved at 0\% noise is poor (showing perhaps that the optimal risk is large or that number of examples is inadequate), the accuracy drops with noise. However, in all cases the drop in accuracy with MAE is much less than that with CCE. This is in accordance with our results on robustness of symmetric losses. We also see that robustness achieved by MSE (which is a bounded loss though it is not symmetric) is in between that of MAE and CCE. As can be seen from the last column of the table, MAE exhibits fair amount of robustness under class conditional noise also. 		
	We observed that higher dropout rates reduce sensitivity of CCE to label noise. This is expected as dropout works like regularizer \cite{srivastava2014dropout}. Interestingly, even without dropout, in many datasets MAE showed good robustness under label noise. On MNIST data with zero dropout, MAE retained almost same accuracy. We did not include these results  in the table because it is customary to have high dropout rate. 
	
\section{Conclusion}
In this paper, we derived some theoretical results on robustness of loss functions in multi-class classification. Such robust loss functions are useful because we can learn a good classifier (without any change in the algorithm or network architecture) even when training set labels are noisy. While we discussed these in the context of learning neural networks, our theoretical results are general and apply to any multi-class classifier learning through risk minimization. For learning neural networks, we showed that the commonly used CCE loss is sensitive to label noise while MAE loss is robust. We presented extensive empirical results to illustrate this. However, training a network under MAE loss would be slow because the gradient can quickly saturate while training. On the other hand, training under CCE is fast. Thus, designing better optimization methods for MAE is an interesting problem for future work. This would allow one to really exploit the robustness properties of MAE (and other such symmetric losses) proved here. 

\bibliography{references}
\end{document}